\documentclass[12pt, fullpage, conference]{IEEEtran}
\IEEEoverridecommandlockouts
\usepackage{cite}
\usepackage{amsmath,amssymb,amsfonts}
\usepackage{algorithmic}
\usepackage{graphicx}
\usepackage{textcomp}
\usepackage{xcolor}
\usepackage{verbatim}
\usepackage{geometry}
\def\BibTeX{{\rm B\kern-.05em{\sc i\kern-.025em b}\kern-.08em
    T\kern-.1667em\lower.7ex\hbox{E}\kern-.125emX}}
    
\begin{document}

\title{An Overview Of 3D Object Detection}

\author{\IEEEauthorblockN{Yilin Wang}
\IEEEauthorblockA{\textit{Computer Science in Multimedia} \\
\textit{University of Alberta}\\
Edmonton, Canada \\
yilin28@alberta.ca}
\and
\IEEEauthorblockN{Jiayi Ye}
\IEEEauthorblockA{\textit{Computer Science in Multimedia} \\
\textit{University of Alberta}\\
Edmonton, Canada \\
jye8@alberta.ca}

}

\maketitle

\begin{abstract}
Point cloud 3D object detection has recently received major attention and becomes an active research topic in 3D computer vision community. However, recognizing 3D objects in LiDAR (Light Detection and Ranging) is still a challenge due to the complexity of point clouds. Objects such as pedestrians, cyclists, or traffic cones are usually represented by quite sparse points, which makes the detection quite complex using only point cloud. In this project, we propose a framework that uses both RGB and point cloud data to perform multiclass object recognition. We use existing 2D detection models to localize the region of interest (ROI) on the RGB image, followed by a pixel mapping strategy in the point cloud, and finally, lift the initial 2D bounding box to 3D space. We use the recently released nuScenes dataset---a large-scale dataset contains many data formats---to training and evaluate our proposed architecture.
\end{abstract}


\section{Introduction}
\label{intro}
The task of target detection is to find all Region of Interests (ROI) in the image and determine their positions and categories. Due to the different appearance, shape and attitude of various objects, as well as the interference of lighting, shielding and other factors during imaging, object detection has continuously been a challenging problem in the field of computer vision.

In this literature review, we summarized some state-of-the-art object detection related works. In section 2, We first briefly describe the data format commonly used in object detection tasks, also pointing out some famous data sets. Then we generalize some proposed methods of preprocessing. In section 3, we introduce techniques related to 2D object detection, including traditional methods and deep learning methods. Finally, we discuss 3D object detection topics in broad outline in section 4.

\section{Data Format}
\subsection{Data Set}
In computer graphics, a Depth Map is an image or image channel that contains information about the distance of the surface of the scene object from the viewpoint. A Depth Map is similar to a grayscale image, except that each pixel is the actual distance between the sensor and the object. In general, RGB images and Depth images are registered, so there is a one-to-one correspondence between pixels.The data set in RGB-D format includes Pascal VOC\cite{everingham2010pascal}, COCO\cite{lin2014microsoft}, ImageNet\cite{5206848}, etc.

Radar data is also useful in object detection problem. Radar data is collected by sending radio waves toward the object surface, then it uses the reflection information to calculate the velocity of objects and the distance to objects. However, radars alone are not able to provide enough information for detection and classification, that is why the fusion of different types of data is very important.

Point cloud data refers to a set of vectors in a three-dimensional coordinate system. These vectors are usually expressed in X,Y, and Z three-dimensional coordinates and are generally used to represent the outer surface shape of an object. Not only that, in addition to the geometric position information represented by (X,Y,Z), each point cloud may include
RGB color pixel, gray value, depth, and normals. Most point cloud data is generated by 3D scanning devices, such as Lidar (2D/3D), STEREO camera, and time-of-flight camera. These devices automatically measure the information of a large number of points on the surface of an object, and then output the point cloud commonly in two standard file format types: .LAS and .LAZ. These point cloud data are collected by scanning equipment. In the data set used for training, there will be corresponding RGB images in addition to the point cloud data. This category of dataset includes KITTI\cite{geiger2013vision}, nuScenes\cite{caesar2020nuscenes}, Waymo Open\cite{sun2019scalability}, etc.

We will use the nuScenes dataset to train and evaluate our model. The nuScenes dataset by nuTonomy ~\cite{nuscenes2019} is a large-scale autonomous driving dataset with 3d object annotations. Compared to many other datasets, nuScenes dataset not only has larger scale with more object
annotations, but also provides entire entire sensor suite, including LIDAR, RADAR, GPS and IMU. Figure 1 presents an example of LIDAR point cloud from nuScenes.

\begin{figure}[htbp]
\centerline{\includegraphics[width=3in]{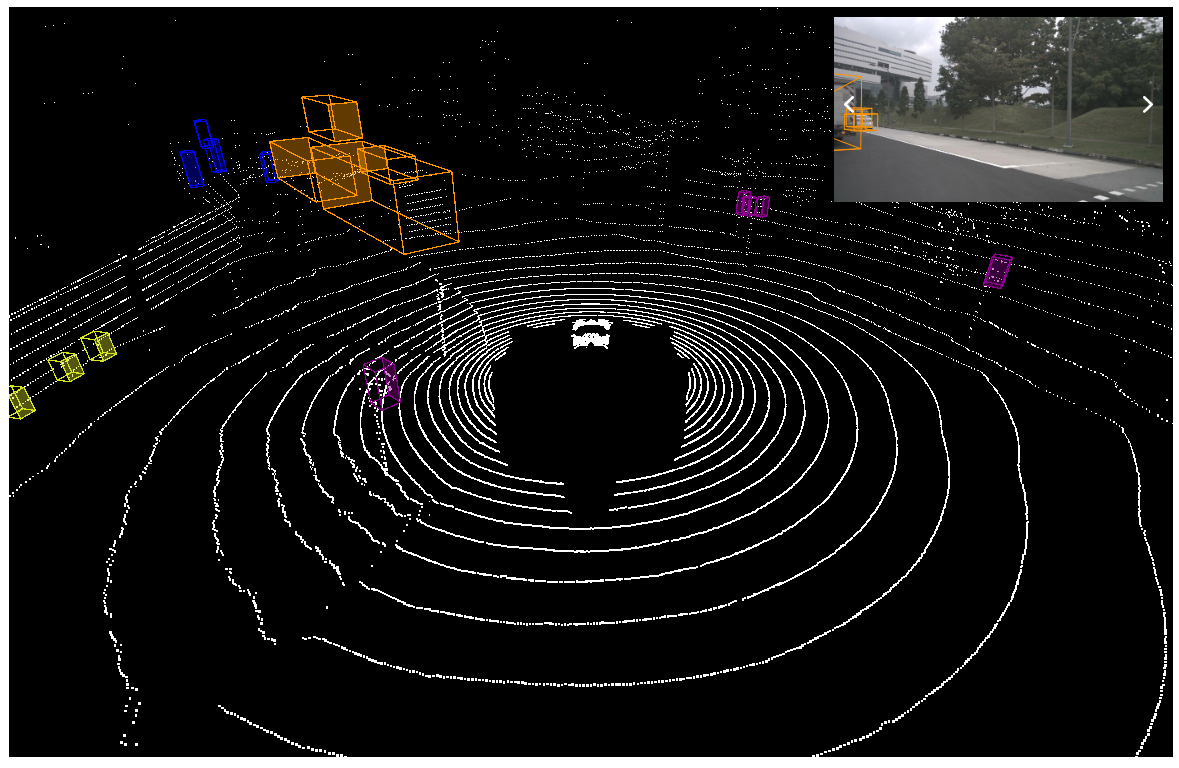}}
\caption{Example of LIDAR point cloud from nuScenes~\cite{nuscenes2019}.}
\label{LIDAR}
\end{figure}

\subsection{Preprocessing}

Although the images in the data set have high quality, in practical application, factors such as weather may affect the quality of the images and thus reduce the detection accuracy. Some recently proposed defogging algorithms for traffic scenes can solve this kind of problem. \cite{liu2019fog} proposed a defogging method based on gamma correction and guided filtering. The method gamma-corrects the image before defogging, and then conducts three different scale guided filtering processes on the gamma corrected image. The filtered image was modified by Retinex model and finally weighted fusion was used to obtain the result of defogging. The defogging picture obtained by this method has higher contrast and color consistency. Wang\cite{wang2019traffic} focused the research on how to solve the problem of uneven distribution of fog concentration. According to the correlation between wavelength and fog concentration, they built a wave-related physical imaging model suitable for traffic monitoring images. Then, according to the correlation between wavelength and color, a transmission estimation algorithm based on maximum fuzzy correlation graph cut is designed.

In addition to improving image quality, camera calibration is often used to determine the position of an image in 3D space. Although camera calibration has always been a widely studied problem, most calibration methods fix the camera position and use some known calibrating patterns. Basu\cite{basu1995active}proposed a method to automatically calibrate the camera by using clear edges in the scene and moving the camera, without defining the pattern in advance.

Apart from the problems mentioned above, some data sets suffer from having serious class imbalance issue, such as the nuScenes Dataset. \cite{zhu2019class} proposed a method to handle this issue effectively and achieves good performance in this new data set. They first use data augmentation strategies, then some improved 3D feature extraction networks, finally
enhance the training procedure and the evaluation of loss to make overall improvement.

\section{2D Object Detection}
\subsection{Traditional Methods}

Traditional target detection algorithms are usually based on various kinds of feature descriptors.
One famous descriptor is the Histogram of Oriented Gradient (HOG)\cite{dalal2005histograms}. It counts occurrences of gradient orientation in localized portions of an image. Hog feature combined with SVM classifier has been widely used in object detection, especially in pedestrian detection with great success. 

Feature detection plays an important role in traditional methods. In recent years, there have been many interesting studies on feature extraction.

Hough transform is one of the basic methods to recognize geometric shapes from images in image processing. For instance, \cite{yin1999integrating} utilizes a
gradient-based Hough transformation to locate the eyes' iris position in a face tracking problem. However, for images with non-Single viewpoint (SVP) criteria, such transform cannot be directly used for feature recognition.\cite{fiala2002hough} put forward one a mathematical model to solve this problem.

Radon Transform \cite{helgason1980radon} is widely used in medical image processing. It can also be used in recognition tasks. In a ground air traffic scenario, \cite{1545408} apply Radon Transformation in visual gesture recognition, result in promising recognition rates.

Yin \cite{yin2001nose} proposed a method to track the shape of the nose, a feature neglected in previous studies. In this method, the area growth method is used to determine the area where the nose is located, and the shape of the tip and the nose wing are extracted respectively through predefined templates. Finally, the extracted features are used to indicate the effect of facial tracking.

Once the interest features are detected, feature trackers like the Kanade–Lucas–Tomasi feature tracker (KLT) can be used to track the features in the next frames. Authors in 2005\cite{Singh2005} proposed an approach utilize Laplacian of Gaussian (LoG) and Gaussian weighting functions to improve the KLT tracking performance, which is subjected to noise. Edge characteristics is coupled into the weighting function, resulting in a deterministic formula for choosing the optimal weighting function. In this way, increase a little computational time while the tracking performance improve tremendously.

Sometimes besides feature extraction, image segmentation is also involved.
\cite{cheng2007airway} describes a methodology that extracts relevant contours on 3D slices by using gradient vector flow (GVF) snakes. The original GVF snake method is modified by adding edge detection and snake shifting techniques using prior knowledge of airway CT slices to gain a better result. This technique may also be useful.

Another problem is induced by the camera motion, as the types of devices collecting data increases, there are more and more data coming from moving cameras. As for the Background Subtraction problem, while many methods work well on data obtained from a still camera, for example, under the prior knowledge that there is a great distinction between foreground and background in each frame, clustering can be done in one iteration and only need two clusters, \cite{xiaom2007} is able to achieve high accuracy in background subtraction. However, dealing with a moving camera is more challenging. This method\cite{bsonmovingcamera} first used the Robust Principal Component Analysis (RPCA) to extract background motion, assuming the background scenes motion can be described as a low-rank matrix, then segment the frames into sub-pixels to enhance the accuracy of converting optical flow into magnitude and angel of motion, thus improve the results.

\subsection{Deep Learning Methods}
Object recognition refers to a collection of related tasks for identifying objects in images or videos. The aim is to find all the objects of interest in the image and determine their 2D location. The region of interest is often defined by a bounding box.

Nowadays, deep learning methods in the field of target detection are mainly divided into two categories: target detection algorithms of two stages; Target detection algorithm of One Stage. The former is a series of candidate boxes generated by the algorithm as samples, and then the samples are classified by the convolutional neural network. The latter can directly transform the target border positioning problem into the regression problem without generating candidate boxes. Because of the difference between the two methods, the performance of the two methods is also different. The former is superior in detection accuracy and positioning accuracy, while the latter is superior in algorithm speed.

Moreover, there are mainly two types of framework in generic. The first one is a two-stage framework. In this kind of framework, region proposals are generate and subsequently classify into different object categories. So it is also called region-based method. This type of model mainly include R-CNN\cite{girshick2014rich}, Fast R-CNN\cite{girshick2015fast}, Faster R-CNN\cite{ren2015faster}, region-based fully convolutional network (R-FCN) \cite{dai2016r}and so on. In single-stage, models regard the object detection task as a unified, end-to-end regression problem. In this kind of framework, images are scaled to the same size and are equally divided into grids. If the center of a target falls in a grid unit, this grid unit is responsible for predicting the target. In this way, the model only needs to process the picture once to get both the location and classification. One-stage frameworks mainly include MultiBox\cite{erhan2014scalable}, YOLO\cite{redmon2016you}, Single Shot MultiBox Detector (SSD)\cite{liu2016ssd}.
Compared with the first type of framework, this kind of framework usually has a simpler structure and a faster detection speed.

\section{3D Object Detection}
In this section, we briefly discussed some of the existing works related to 3D object detection. 
We divided those works into three main categories based on different data information. 

\subsection{Object detection using RGB image}
\label{rgb only}
\begin{figure}[htbp]
\centerline{\includegraphics[width=3in]{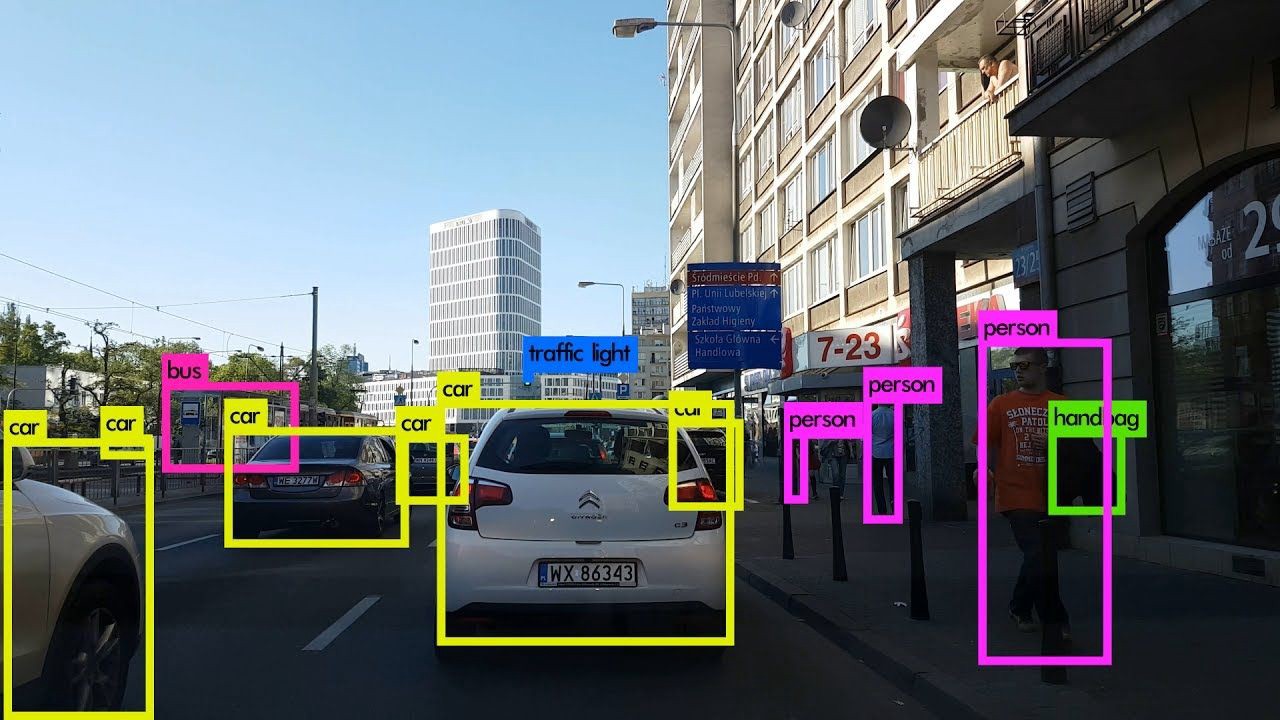}}
\caption{Example of object detection in a 2D image.}
\label{2Ddetection}
\end{figure}

RGB images are very useful for object detection for their sufficient semantic information.
\ref{2Ddetection} is an object detection example using 2D image. Methods like 3D-GCK~\cite{3dgck} uses monocular RGB images exclusively and can achieve real-time vehicles detection. It first predicts 2D bounding boxes, and uses neural networks to estimate 
the missing depth information, lifting 2D bounding boxes to 3D space.

A recent work\cite{Lugo2020} uses an RGB-D sensor, yet utilize only the grayscale information to recognize textureless objects. It first converts the RGB images obtained from the sensor to gray scale images, then segment the background and the foreground. After moving the noise, five classification models are applied to do the feature extraction and eventually predict the class of the objects.

\subsection{Object detection using Point Cloud}
\label{lidar only}
\begin{figure}[htbp]
\centerline{\includegraphics[width=3in]{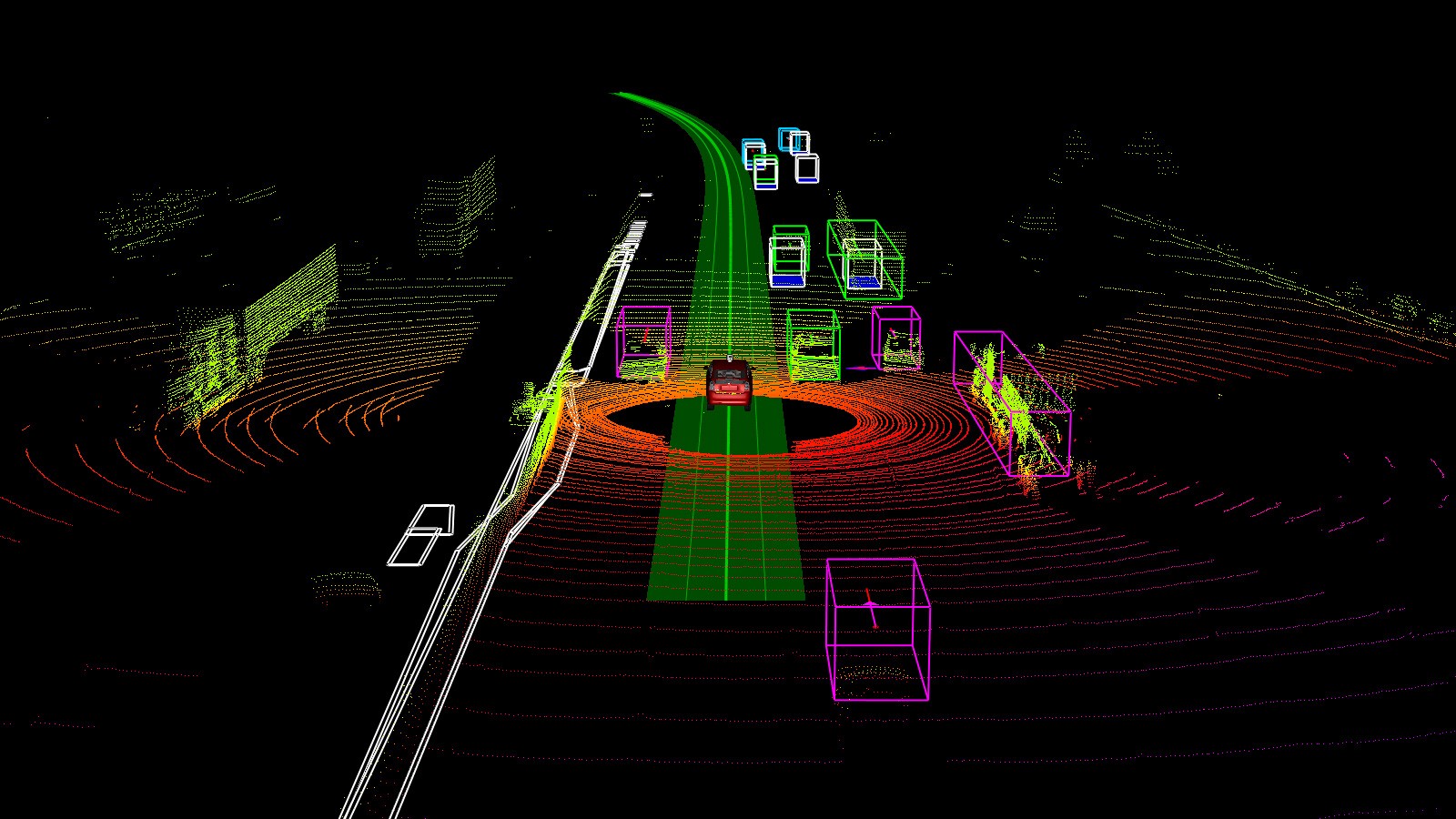}}
\caption{Example of object detection in 3D Point Cloud from by YOLO.}
\label{3DLIDAR}
\end{figure}

There are mainly two types of classification networks that only use point cloud data. The first type of method directly uses three-dimensional point cloud data. This type of method does not suffer from information loss, but it is often computationally expensive due to the complexity of 3D data. The second type of method reduces the amount of computation by processing the point cloud into two-dimensional data, but inevitably loses some features of the original data. Figure \ref{3DLIDAR} shows an example of detection in 3D lidar point cloud data.

There are many ways to process a point cloud into two dimensional data.
Jansen's\cite{jansen20193d} approach is to represent 3D Point Clouds by using several 2D perspectives. In the data preprocessing stage, the point cloud image is intercepted from 64 different angles by taking the center of the image as the origin and rotating with a fixed radius. This method reduces information loss to a certain extent since additional rotation is added to the scene and many instances are utilized during the classification process. Others like \cite{li2016vehicle} project the point cloud on a cylinder plane to reserve as much information as possible.

BirdNet+~\cite{birdnetplus} is an improvement of BirdNet~\cite{inproceedings}, an object detection framework. BirdNet+ provides an end-to-end solution for 3D object detection on LiDAR data. It takes the Bird Eyes View representations, which are 2D structures with three channels that are transformed from LiDAR point cloud, relies on a two-stage architecture to obtain the 3D oriented boxes.

Recently, \cite{Chen202007} presented a computational efficiency end-to-end robust point cloud alignment and object recognition method using unsupervised deep learning, named Deep Point Cloud Mapping Network (DPC-MN). This model can be trained without labels and efficiently implement the mapping function to 2D view from it 3D point cloud representation.

\subsection{Object detection combined RGB images with Point Cloud}
\label{rgb with lidar}
Frustum PointNets\cite{qi2018frustum} use both RGB images and LiDAR point cloud data for 3D object detection. This algorithm takes the advantage of mature 2D object detectors to reduce the search space.
It extracts the 3D bounding frustum of an object by extruding 2D bounding boxes from image detectors and then perform 3D object instance segmentation in the 3D space trimmed by the 3D frustums.

\begin{figure}[htbp]
\centerline{\includegraphics[width=3in]{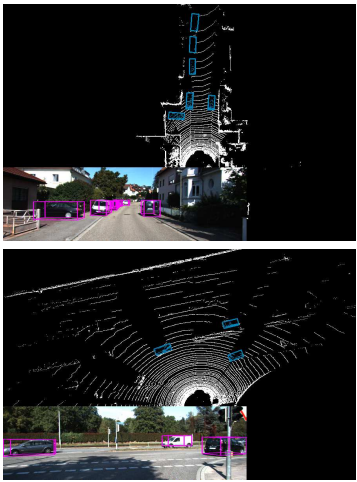}}
\caption{Example of object detection by MV3D.}
\label{MV3D}
\end{figure}

MV3D\cite{chen2017multi} also uses both RGB images and LiDAR point cloud data as input. It projects the 3D point cloud to the bird’s eye view and the front view. The bird’s eye view representation is encoded by height, intensity and density, while the front view is generated by projecting the point cloud to a cylinder plane. The bird's eye view is used to generate 3D prior box, the 3D prior box is then projected to the front view and the image. All three inputs generate a feature map. ROI pooling is adopted to integrate the three feature map into the same dimension. The integrated data is fused over the network, and output classification results and bounding box.

\section{References}


\bibliographystyle{IEEEtran}
\bibliography{egbib}

\end{document}